\documentclass[10pt,twocolumn,letterpaper]{article}

\usepackage{cvpr}              
\usepackage{multirow}

\usepackage[dvipsnames]{xcolor}

\definecolor{cvprblue}{rgb}{0.21,0.49,0.74}
\usepackage[pagebackref,breaklinks,colorlinks,citecolor=cvprblue]{hyperref}

\def\paperID{xxx} 
\def\confName{xxx\xspace}
\def\confYear{xxxx\xspace}

\title{DEGAS: Detailed Expressions on Full-Body Gaussian Avatars}

\author{
Zhijing Shao$^{1,2}$ \qquad
Duotun Wang$^{1}$ \qquad
Qing-Yao Tian$^{2}$ \qquad
Yao-Dong Yang$^{1}$ \qquad
Hengyu Meng$^{1}$ \qquad
\\
Zeyu Cai$^{1}$ \qquad
Bo Dong$^{4}$ \qquad
Yu Zhang$^{2}$ \qquad
Kang Zhang$^{1,3}$ \qquad
Zeyu Wang$^{1,3}$ \qquad
\\
$^{1}$The Hong Kong University of Science and Technology (Guangzhou)
\\
$^{2}$Prometheus Vision Technology Co., Ltd.
\\
$^{3}$The Hong Kong University of Science and Technology
\\
$^{4}$Swinburne University of Technology
\\
}

\begin{document}

\twocolumn[{
\maketitle
\begin{center}
    \captionsetup{type=figure}
    \includegraphics[width=\textwidth]{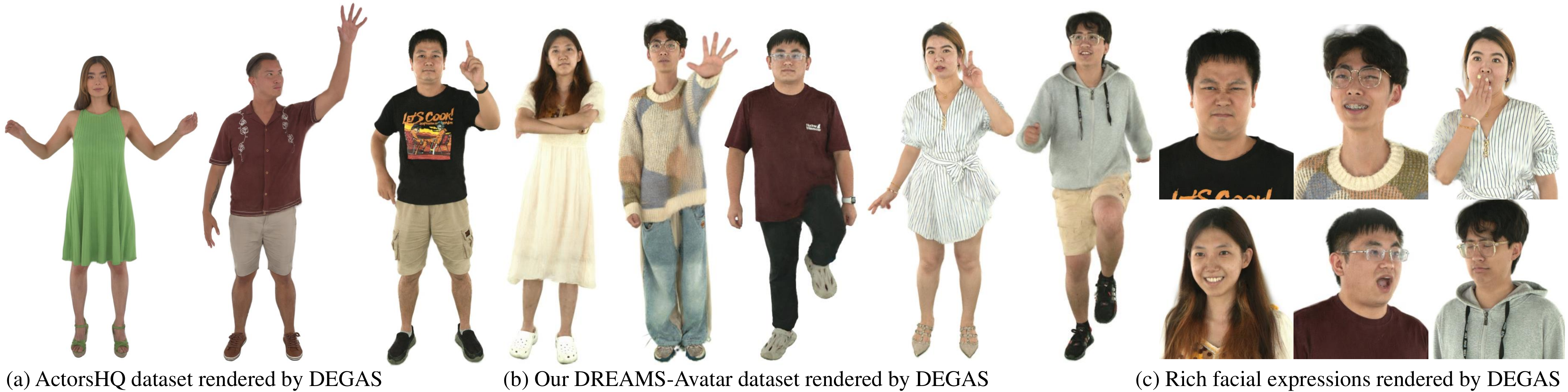}
    \captionof{figure}{
    \textbf{Photorealistic rendering of full-body avatars using our method on (a) ActorsHQ dataset and (b) our proposed DREAMS-Avatar dataset, (c) with rich facial expressions.}
    }
    \label{fig:teaser}
\end{center}
}]

\begin{abstract}
Although neural rendering has made significant advances in creating lifelike, animatable full-body and head avatars, incorporating detailed expressions into full-body avatars remains largely unexplored.
We present DEGAS, the first 3D Gaussian Splatting~(3DGS)-based modeling method for full-body avatars with rich facial expressions.
Trained on multiview videos of a given subject, our method learns a conditional variational autoencoder that takes both the body motion and facial expression as driving signals to generate Gaussian maps in the UV layout.
To drive the facial expressions, instead of the commonly used 3D Morphable Models (3DMMs) in 3D head avatars, we propose to adopt the expression latent space trained solely on 2D portrait images, bridging the gap between 2D talking faces and 3D avatars. 
Leveraging the rendering capability of 3DGS and the rich expressiveness of the expression latent space, the learned avatars can be reenacted to reproduce photorealistic rendering images with subtle and accurate facial expressions.
Experiments on an existing dataset and our newly proposed dataset of full-body talking avatars demonstrate the efficacy of our method. 
We also propose an audio-driven extension of our method with the help of 2D talking faces, opening new possibilities for interactive AI agents.
Project page:
\url{https://initialneil.github.io/DEGAS}
.
\end{abstract}
\vspace{-0.5em}

\begin{figure*}[htbp]
    \centering
    \includegraphics[width=\linewidth]{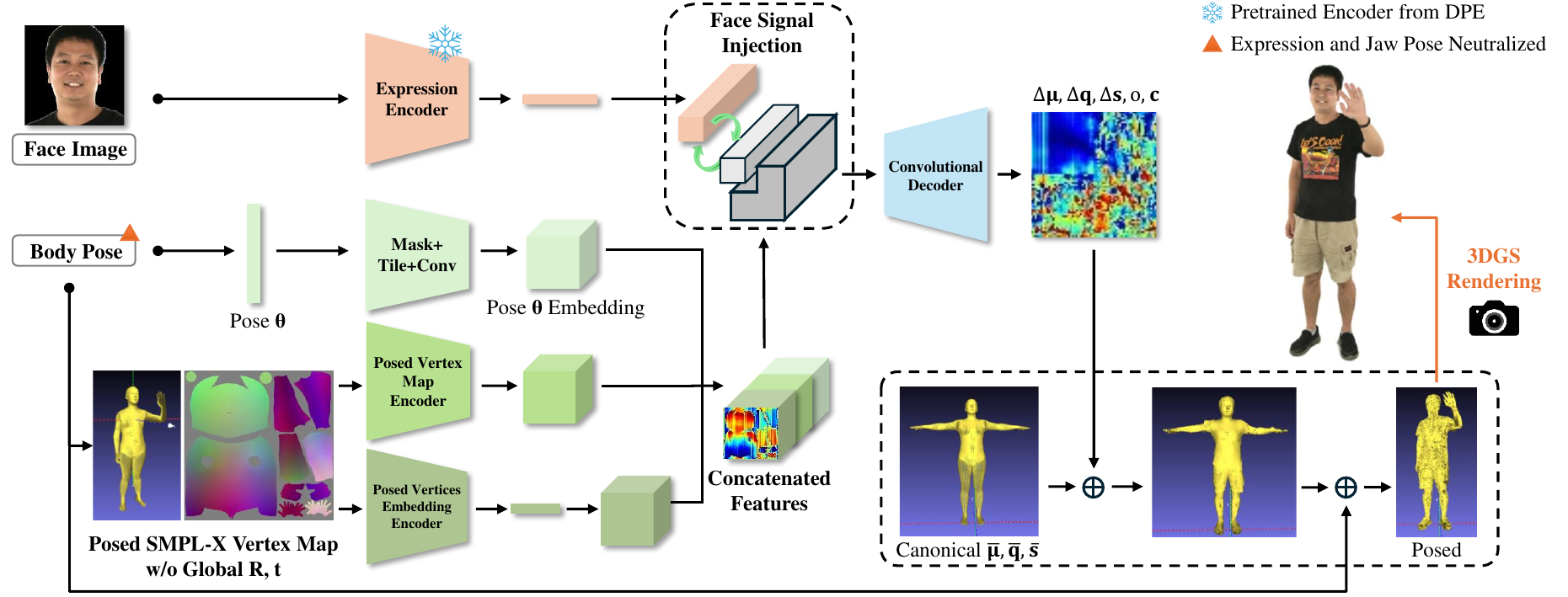}
    \caption{ 
    \textbf{The pipeline of our method.} 
    DEGAS takes face signal from the pretrained expression encoder of DPE~\cite{DPE:CVPR:2023}, which is injected to the body signal from SMPL-X~\cite{SMPL-X:CVPR:2019}.
    The pose-dependent Gaussian maps generated by the convolutional decoder are applied to the pose-independent maps for 3DGS~\cite{3DGS:SIGGRAPH:2023} rendering.
    }
    \label{fig:framework}
\end{figure*}

\section{Introduction}
\label{sec:intro}

Photorealistic and animatable human modeling has been an active research topic in computer vision and graphics for decades.
Interactive avatars that are capable of performing natural body motions and subtle facial expressions can benefit numerous downstream applications, e.g., tele-presentation~\cite{VTour:IEEEVR:2020, VMirror:CHI:2021}, virtual companion~\cite{bevacqua2010greta}, and extended reality (XR) storytelling~\cite{ARBook:CE:2014, MRBook:ISMAR:2021}.

With the rise of neural rendering such as Neural Radiance Fields~(NeRF)~\cite{NeRF:ECCV:2020} and 3DGS~\cite{3DGS:SIGGRAPH:2023}, we observe a boost in terms of quality and rendering efficiency for both \textit{full-body avatars}~\cite{jiang2023instantavatar, li2024animatablegaussians, hu2024gaussianavatar} and \textit{head avatars}~\cite{zielonka2023insta, shao2024splattingavatar, GaussianAvatars:CVPR2024, RGCA:CVPR2024}.
Yet there is a lack of dataset and method for integrating the \textit{two}, i.e., expressive full-body avatars equipped with both body pose control and rich facial expressions.
We aim to fill in the gap by proposing
\textbf{DEGAS}~(\textbf{D}etailed \textbf{E}xpressions on full-body \textbf{G}aussian \textbf{A}vatar\textbf{s}),
the first 3DGS-based method for deling full-body talking avatars together with a new dataset for evaluation.

A naive way to enable facial expressions on full-body avatars would be using 3DMMs for facial control, for example, controlling the \textit{expression} parameters of SMPL-X~\cite{SMPL-X:CVPR:2019}, or using the parameters of FLAME~\cite{FLAME:SiggraphAsia2017} or BFM~\cite{BFM:2009} as driving signals.
We find two drawbacks of such naive methods.
1)~The expressiveness of 3DMMs is limited~\cite{DPE:CVPR:2023, xing2023codetalker, HeadEvolver:arXiv:2024}. 
Being coarse meshes, 
3DMMs are essentially not able to capture subtle facial changes.
2)~Driving signal generation for a 3DMM is non-trivial. 
Both video-driven~\cite{MICA:ECCV2022, DECA:Siggraph2021, dad3dheads:CVPR:2022} and audio-driven~\cite{faceformer:CVPR:2022, xing2023codetalker, Imitator:ICCV:2023, peng2023synctalk, FaceTalk:CVPR:2024} methods suffer from efficiency, accuracy, and expressiveness issues.
Recent works~\cite{GaussianAvatars:CVPR2024, ng2024audio2photoreal} have achieved photorealistic facial modeling in a very dense multi-view camera set-up. However, accurate 3DMM registration is non-trivial from a monocular camera~\cite{DECA:Siggraph2021, dad3dheads:CVPR:2022}. 

One important trend we observe in recent works on 2D talking faces is the learning of disentangled latent spaces for identity, pose, and expression~\cite{wang2021facevid2vid, drobyshev2022megaportraits, DPE:CVPR:2023, xu2024vasa}. 
Such a framework enables standalone extraction of identity-agnostic pose and expression parameters from input images. 
We propose to adopt the pretrained encoder for expression from DPE~\cite{DPE:CVPR:2023} to generate driving signals for the facial control of our avatar modeling.
To the best of our knowledge, we are the first method to bridge the gap between 2D talking faces and 3D avatars.
We take inspiration from 
the Score Distillation Sampling proposed in DreamFusion~\cite{poole2022dreamfusion} where a pretrained 2D generative model lays the foundation for 3D generative tasks.

One of the benefits of involving a pretrained encoder is that we can extend audio-driven 2D talking faces to our 3D avatars. 
Given one portrait image and an audio clip, we first use SadTalker~\cite{Sadtalker:CVPR:2023} to generate the corresponding talking head video, and then apply the pretrained expression encoder to extract driving signals for our avatars. 


Though both trained with registered meshes, head avatars and full-body avatars usually face very different registration qualities.
With well aligned mesh surfaces,
the binding of the 3D Gaussians and the underlying mesh is usually simple for head avatars. 
Both SplattingAvatar~\cite{shao2024splattingavatar} and GaussianAvatars~\cite{GaussianAvatars:CVPR2024} propose to bind 3D Gaussians to FLAME mesh triangles without any pose-dependent compensations. 
Full-body avatars, on the other hand, usually face the challenge of a much less accurate underlying mesh because of clothing.
D3GA~\cite{Zielonka2023Drivable3D} proposes to alleviate this problem by modeling clothes with separate tetrahedron layers.
We follow the practice of AnimatableGaussians~\cite{li2024animatablegaussians} and CodecAvatars~\cite{bagautdinov2021drivingsignal} to leverage the ability of 2D CNNs for the pose-dependent generation of 3DGS parameters.


\vspace{0.5em}
\noindent 
\vspace{0.5em}
In summary, our main contributions are as follows:
\begin{itemize}[leftmargin=1.25em]
\setlength{\itemsep}{0.25em}
\setlength{\parsep}{0.25em}
\setlength{\parskip}{0.25em}

\item We propose the first 3DGS-based method for full-body talking avatars and a multi-view captured dataset of full-body avatars with rich facial expressions.

\item We propose to drive 3D avatars with 2D talking faces, bridging the gap between these two research topics and opening new possibilities for the reenactment of photorealistic avatars.

\end{itemize}

\section{Related Work}
\label{sec:related}

\subsection{3D Avatar Representations}
3D avatar modeling methods have three major design choices to make: appearance representation, canonical modeling, and posing method.
The change of appearance representation from mesh texture~\cite{habermann2021ddc, pixel_codec_avatars, bagautdinov2021drivingsignal} to points~\cite{zheng2023pointavatar}, NeRF~\cite{jiang2023instantavatar, zielonka2023insta}, and 3DGS~\cite{shao2024splattingavatar, GaussianAvatars:CVPR2024, li2024animatablegaussians, hu2024gaussianavatar}, has been the driving force of quality improvements in this field.

In terms of canonical modeling, there are two major categories depending on whether the appearance is pixel-wisely defined on the UV space or not.
UV atlas of SMPL~\cite{SMPL:TOG:2015}, SMPL-X~\cite{SMPL-X:CVPR:2019}, and FLAME~\cite{FLAME:SiggraphAsia2017} provides a well aligned layout for pixel-wise appearance representation.
CodecAvatars~\cite{pixel_codec_avatars, bagautdinov2021drivingsignal, ng2024audio2photoreal}, HDHumans~\cite{HDHumans}, DDC~\cite{habermann2021ddc}, and UVVolumes~\cite{chen2023uv}
predict pixel-wise appearance features on the UV space.
RGCA~\cite{RGCA:CVPR2024}, ASH~\cite{pang2024ash}, and GaussianAvatar~\cite{hu2024gaussianavatar}
construct pixel-wise 3D Gaussian parameters.
AnimatableGaussians~\cite{li2024animatablegaussians} further uses front and back planes to utilize the geometry details from the mesh.

When the UV layout is not used, the canonical model is usually defined tightly aligned to the underlying mesh.
NeuralBody~\cite{peng2021neural, peng2023implicit} encodes learnable latent codes to SMPL mesh vertices.
EditableHumans~\cite{locally_editable_humans} assigns a learnable codebook to the vertices of SMPL-X.
SLRF~\cite{zheng2022structured} learns structured local radiance fields attached to SMPL.
AvatarRex~\cite{zheng2023avatarrex} models the appearance with feature planes aligned to the canonical mesh.
TAVA~\cite{li2022tava}, TotalSelfScan~\cite{dong2022totalselfscan}, PoseVocab~\cite{li2023posevocab}, InstantAvatar~\cite{jiang2023instantavatar}, and INSTA~\cite{zielonka2023insta} propose to build NeRF aligned with the canonical mesh.
PointAvatar~\cite{zheng2023pointavatar} constructs canonical points coupled with FLAME canonical space.
ARAH~\cite{ARAH:ECCV:2022} and X-Avatar~\cite{xavatar} define pose-conditioned colors on the surface of the canonical SDF.
SplattingAvatar~\cite{shao2024splattingavatar} and GaussianAvatars~\cite{GaussianAvatars:CVPR2024} learn 3D Gaussians embedded on the underlying mesh triangles.
3DGS-Avatar~\cite{qian20233dgsavatar} and GauHuman~\cite{hu2024gauhuman} initialize 3D Gaussians by sampling from SMPL.
To take advantage of powerful 2D CNNs, our method falls in the former category with pixel-wise 3DGS parameters defined 
as UV maps.

With the appearance and canonical modeling chosen, the posing scheme helps warp sample points from the canonical space to the posed space or vice versa depending on the need for rendering.
Posed-dependent compensation is usually introduced together with LBS to achieve higher quality~\cite{pixel_codec_avatars, bagautdinov2021drivingsignal, RGCA:CVPR2024, hu2024gaussianavatar, li2024animatablegaussians},
while direct posing from mesh triangles increases the inference FPS~\cite{zielonka2023insta, shao2024splattingavatar, GaussianAvatars:CVPR2024}.
Appearance modeling methods of texture, points, and 3DGS favor a forward posing scheme, i.e., per-primitive conversion from the canonical space to the posed space. NeRF-based methods~\cite{li2022tava, dong2022totalselfscan, jiang2023instantavatar, li2023posevocab}, on the other hand, usually require the conversion of multiple samples on each ray from the posed space back to the canonical space~\cite{snarf, fast-snarf}.

\subsection{Talking Face Video Generation}
With rapid advances in GANs and diffusion models, talking-face video generation has achieved remarkable quality improvements. For instance, view-dependent landmarks~\cite{FewLmk:NIPS:2019} and 3DMMs~\cite{yin2022styleheat} have been leveraged to enhance the temporal consistency of synthesized facial animations. Recent investigations have focused on distilling disentangled information such as emotion~\cite{EmotionDis:arXiv:2024, FaceChain:CVPR:2024}, expression~\cite{DPE:CVPR:2023}, and appearance~\cite{MTalk:arXiv:2024}, enabling face reenactment and editing across identities. 
For producing different speaking styles, image translation approaches~\cite{ImgTrans:CVPR:2017, GAN:NIPS:2014} are utilized (e.g., HeadGAN~\cite{Headgan:NIPS:2021} and SadTalker~\cite{Sadtalker:CVPR:2023}). 
To express a realistic appearance with natural pose modifications, researchers have extended NeRF and 3DGS to talking face tasks, as seen in GaussianTalk~\cite{Gaussiantalker:arXiv:2024} and Ad-NeRF~\cite{Ad-NERF:CVPR:2021}.




\subsection{Choice of Driving Signal}
Different from a 4D playback system like FVV~\cite{fvv}, or 4DGS~\cite{wu20244dgs}, avatar modeling methods focus on the reenactment from accessible driving signals.
For full-body avatars, apart from the commonly used skeleton poses as driving signals~\cite{qian20233dgsavatar, jiang2023instantavatar, hu2024gaussianavatar}, AnimatableGaussians~\cite{li2024animatablegaussians} also encodes the posed position maps.
SurMo~\cite{hu2024surmo} further considers temporal dynamics to overcome the limitation of the deterministic nature of driving signals.
Head avatars~\cite{zielonka2023insta, zheng2023pointavatar, GaussianAvatars:CVPR2024, shao2024splattingavatar}, on the other hand, usually take into account the expression parameters of the underlying 3DMMs for better capturing the surface deformation of the face mesh.
Audio2Photoreal~\cite{ng2024audio2photoreal} models mesh-based full-body avatars with facial control of accurate expression parameters registered from a dense camera set-up, which is non-trivial to acquire from a more casual set-up.

Recent works in talking face video generation have explored different driving signals for both audio-driven and video-driven tasks.
Some~\cite{ren2021pirenderer, yin2022styleheat, peng2023synctalk} follow the practice of 3D face animation~\cite{VOCA2019, faceformer:CVPR:2022, xing2023codetalker, Imitator:ICCV:2023} to use parameters from a pretrained 3DMM as the driving signal.
However, the expressiveness of 3DMM is limited.
CodeTalker~\cite{xing2023codetalker} proposes to regress vertex offsets instead of FLAME parameters to improve the expressiveness of face animation.
DPE~\cite{DPE:CVPR:2023}, instead of using 3DMMs, designs a bidirectional cyclic training strategy to construct disentangled latent spaces for pose and expression.
VASA-1~\cite{xu2024vasa} and Hallo~\cite{Hallo:arXiv:2024} inherit this idea and achieve outstanding quality with the diffusion model. In this paper, we aim to discuss that a well-trained latent space of expression solely from 2D images, is a better choice than 3DMM as the driving signal to reenact expressive 3D avatars.

\section{Methodology}
\label{sec:method}

Given synchronized multiview videos and per-frame registered SMPL-X of a given subject, we train an expressive full-body avatar modeled by a conditional variational autoencoder (cVAE) to generate the 3D Gaussian maps in the layout of SMPL-X's UV space, where each pixel parameterizes one 3D Gaussian primitive.
We briefly introduce 3DGS in \Cref{sec:3dgs}, and elaborate on the choice of driving signals in \Cref{sec:driving_signal}, the design of cVAE in \Cref{sec:cVAE}, the LBS-based posing scheme in \Cref{sec:gsmaps_lbs}, and finally the training process in \Cref{sec:training}.

\subsection{Preliminaries: 3D Gaussian Splatting}
\label{sec:3dgs}

3DGS~\cite{3DGS:SIGGRAPH:2023} is an explicit primitive-based 3D representation that models a scene or an object by a set of semi-transparent ellipsoids as 3D Gaussians. 
Each 3D Gaussian has a parameter set of $G_i=(\boldsymbol{\mu}_i, \boldsymbol{q}_i, \boldsymbol{s}_i, o_i, \boldsymbol{\eta}_i)$.
The probability density of each Gaussian in space is formulated by its mean (position) $\boldsymbol{\mu}_i$ and covariance matrix $\Sigma_i$ as:
\begin{equation}
    f(x|\boldsymbol{\mu}_i, \Sigma_i) = e^{- \frac{1}{2} (x-\boldsymbol{\mu}_i)^T \Sigma_{i}^{-1} (x-\boldsymbol{\mu}_i)}
\end{equation}
\begin{equation}
    \Sigma_{i} = R_{i} S_{i} S_{i}^T R_{i}^T
\end{equation}
where the rotation matrix $R_i$ and scaling matrix $S_i$ that formulate the covariance matrix $\Sigma_i$ are constructed from the rotation quaternion $\boldsymbol{q}_i$ and the scaling vector $\boldsymbol{s}_i$ respectively.

Given the world-to-camera view matrix $W$ and the Jacobian $J$ of the point projection matrix, the influence of the 3D Gaussians can be splatted onto 2D~\cite{SurfaceSplatting}:
\begin{equation}
    \Sigma'_i = J W \Sigma_i W^T J^T
\end{equation}
The final rendered color $\boldsymbol{C}$ of a pixel is given by the \emph{$\alpha$-blending} of the 3D Gaussians that splatted onto it from near to far:
\begin{equation}
    \boldsymbol{C} = \sum_{i=1}^{N} {\boldsymbol{c_i}} \alpha_i \prod_{j=1}^{i-1}(1 - \alpha_j)
    \label{blending}
\end{equation}
with $\alpha_i$ evaluated from the splatted covariance $\Sigma'_i$, and the opacity in logit $o_i$ with \text{sigm()} being the standard sigmoid function:
\begin{equation}
    \alpha_i(x) = \text{sigm}(o_i) \exp( -\frac{1}{2} (x - \mu_i) (\Sigma'_i)^{-1} (x - \mu_i) )
\end{equation}
and $\boldsymbol{c_i}$ is the view-dependent color represented by spherical harmonics $\boldsymbol{\eta}_i$.
For simplicity in this paper, we disable the view-dependent components of $\boldsymbol{\eta}_i$ by predicting $\boldsymbol{c_i}$ directly.

\begin{figure}
    \centering
    \includegraphics[width=\linewidth]{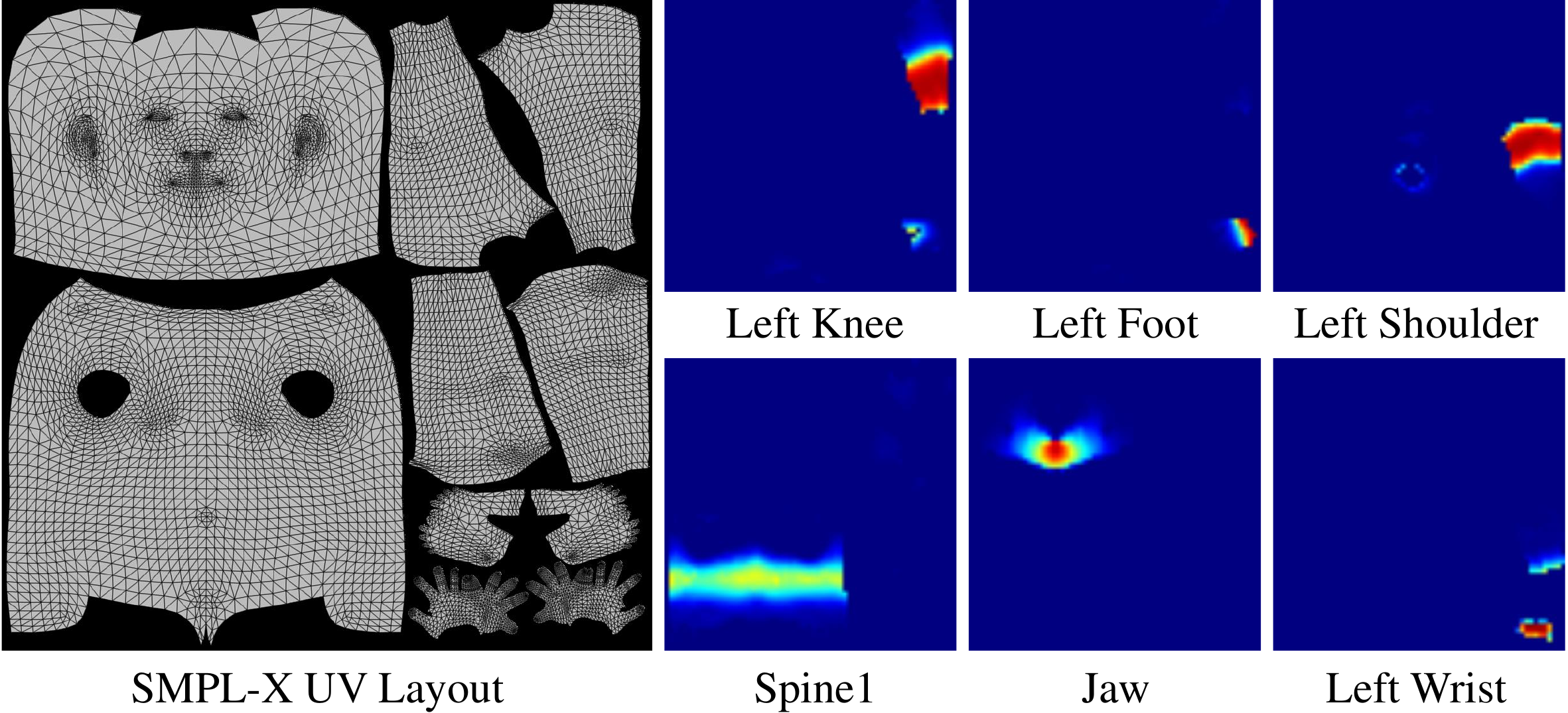}
    \caption{
        \textbf{Tiling and masking of Pose $\boldsymbol\theta$ Embedding.}
        Joint angles are filled to corresponding areas in the UV layout where the joints can affect through skinning. Colors visualize the skinning weights, which are converted to binary masks.
    }
    \label{fig:theta_embs}
\end{figure}

\subsection{Driving Signal}
\label{sec:driving_signal}

For full-body talking avatars, both facial expressions and body gestures convey important messages in the process of communication.
We divide the driving signals into two parts: body signal and face signal.

\noindent
\textbf{Body signal.}
We use the body pose parameters from SMPL-X as the body signal, which enables body motion control down to the finger level.
We keep the motion of the head to make the avatar more natural.
The jaw pose, on the other hand, is neutralized, leaving full facial control to the face signal.

\noindent
\textbf{Face signal.}
Instead of using parameters from a 3DMM, we adopt the pretrained expression encoder from DPE~\cite{DPE:CVPR:2023} to extract pose-agnostic expressions from portrait images.
Unlike 3DMMs~\cite{BFM:2009, FLAME:SiggraphAsia2017} trained from untextured scan meshes, the latent space that 2D talking faces methods like DPE~\cite{DPE:CVPR:2023} constructed from a large number of images is more expressive to capture the expression-related appearance variations.

\begin{figure}
    \centering
    \includegraphics[width=\linewidth]{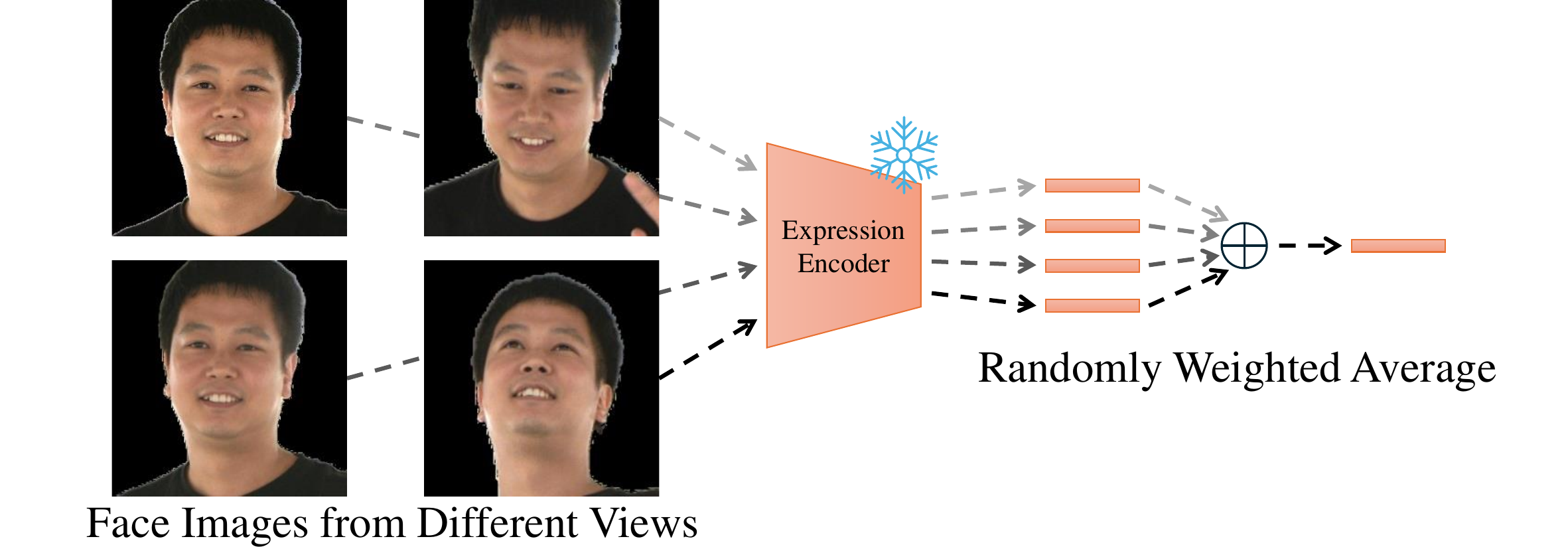}
    \caption{
        \textbf{Training with faces from multiviews.}
        During training, we extract expressions from multiple views and use a randomly averaged latent code as the face signal.
    }
    \label{fig:multi_faces}
\end{figure}

\subsection{Conditional Variational Autoencoder}
\label{sec:cVAE}

Given synchronized multiview images and registered SMPL-X of one frame, we neutralize the jaw pose and expression parameters and encode the body pose $\boldsymbol\theta$ with three encoders.
Injected with a face signal from the expression encoder, the mixed driving signal is fed to a convolutional decoder for the generation of Gaussian parameters.
\Cref{fig:framework} illustrates the pipeline.

\noindent
\textbf{Pose $\boldsymbol\theta$ Embedding.}
The first encoder takes as input the pose vector $\boldsymbol\theta \in \mathbb{R}^{162}$ from SMPL-X with 54 joint angles, excluding the root joint for global orientation. 
We follow the practice of CABody~\cite{bagautdinov2021drivingsignal} to expand the vector by \text{64$\times$64} and then mask by downsized skinning maps from UV.
This encoding scheme sets each pose component to the UV layout only where its corresponding joint affects.
We visualize where the joints' $\boldsymbol\theta$ are filled to in \Cref{fig:theta_embs}.

\noindent
\textbf{Posed Vertex Map Encoders.}
Another two encoders process the posed vertex map on UV.
The first convolution encoder, referred to as the Posed Vertex Map (PVM) Encoder, encodes the posed vertex map down to resolution \text{64$\times$64}, matching that of the pose $\boldsymbol\theta$ encoding.
The second, referred to as the Posed Vertices Embedding (PVE) Encoder, encodes the posed vertex map as a latent code, which we believe better captures the global information of the pose.
The encoded features from the above three branches are concatenated as the final pose feature.

\noindent
\textbf{Expression Encoder.}
DPE~\cite{DPE:CVPR:2023} proposes a bidirectional
cyclic training strategy in the training process aiming at the disentanglement of pose and expression.
We take the pretrained expression encoder as our facial controller.
From the input multiview images, we apply the head pose estimator from SynergyNet~\cite{wu2021synergy} to find front-facing views.
In the training stage, we choose the four most frontal views to extract expression codes. For every iteration, a randomly weighted average of the four codes is used as the face signal, as illustrated in \Cref{fig:multi_faces}.

\noindent
\textbf{Face Signal Injection.}
The expression code is firstly reformed by a fully connected layer and a small convolutional decoder to the resolution of \text{32$\times$32}. Then it is used to replace the top-left quarter of the pose feature.
In the UV layout of SMPL-X, the top-left quarter corresponds to the face region.
After the injection, we feed the mixed feature to a convolutional decoder for extracting Gaussian maps with pose-dependent 
corrections for
position $\Delta\boldsymbol{\mu}$, rotation $\Delta\boldsymbol{q}$, and scaling $\Delta\boldsymbol{s}$, as well as pose-independent opacity $o$, and color $\boldsymbol{c}$.

\subsection{Gaussian Maps and LBS}
\label{sec:gsmaps_lbs}

We define the base Gaussian Maps in SMPL-X's UV layout.
The SMPL-X mesh in T-Pose is rasterized to UV as the base positions $\overline{\boldsymbol{\mu}}$, i.e., every valid pixel on UV represents one 3D Gaussian primitive in the canonical space.
Similar to D3GA~\cite{Zielonka2023Drivable3D}, we find every Gaussian primitive's corresponding triangle on the mesh to initialize its base rotation $\overline{\boldsymbol{q}}$, such that each Gaussian will have its first row axes aligned with the mesh surface and the third with the normal.
The base scaling $\overline{\boldsymbol{s}}$ is initialized by treating the base positions as a regular point cloud.

Given the barycentric-interpolated skinning weights $\mathcal{W}$ from SMPL-X mesh, we apply the pose-dependent corrections in the canonical space, and calculate the Gaussian primitives' posed position $\boldsymbol{\mu}_p$ and rotation $\boldsymbol{q}_p$ by LBS:

\begin{equation}
    \boldsymbol{\mu}_p = LBS_{Rt}(\overline{\boldsymbol{\mu}} + \Delta\boldsymbol{\mu}, \boldsymbol\theta, \mathcal{W})
\end{equation}
\begin{equation}
    \boldsymbol{q}_p = LBS_{R}(\overline{\boldsymbol{q}} + \Delta\boldsymbol{q}, \boldsymbol\theta, \mathcal{W})
\end{equation}
\begin{equation}
    \boldsymbol{s} = \overline{\boldsymbol{s}} + \Delta\boldsymbol{s}
\end{equation}

As in the 3DGS rendering pipeline, the Gaussian primitives in the posed space $G=(\boldsymbol{\mu}_p, \boldsymbol{q}_p, \boldsymbol{s}, o, \boldsymbol{c})$ are splatted onto a 2D image as the final rendering.

\subsection{Loss and Training}
\label{sec:training}

In the training process, DEGAS is supervised with photometric loss of $\mathcal{L}_1$, $\mathcal{L}_{ssim}$, and $\mathcal{L}_{lpips}$~\cite{zhang2018perceptual}.
The pose-dependent position $\Delta\boldsymbol{\mu}$ is regularized by an offset loss.
To help convergency, we also introduce a regularization $\mathcal{L}_{s}$ on scaling to prevent any Gaussians from becoming too large (10x larger than its base scaling):

\begin{equation}
\begin{split}
    \mathcal{L} = 
    (1-\lambda_{ssim}) \mathcal{L}_1 
    + \lambda_{ssim} \mathcal{L}_{ssim}
    + \lambda_{lpips} \mathcal{L}_{lpips}
    \\
    + \lambda_{\boldsymbol{\mu}} 
    \left\| \Delta\boldsymbol{\mu} \right\| _2
    + \lambda_{s} \mathcal{L}_{s}
\end{split}
\end{equation}
\begin{equation}
    \mathcal{L}_{s} (i) =  
    \left\{
     \begin{array}{lr}
     \vert \boldsymbol{s}_i \vert, &  \boldsymbol{s}_i > 10 \overline{\boldsymbol{s}}_i \\
     0, & otherwise \\
     \end{array}
    \right.
\end{equation}
where $\lambda_{ssim} = 0.2$, $\lambda_{lpips} = 0.1$, $\lambda_{\boldsymbol{\mu}} = 0.001$, and $\lambda_s = 1.0$ all through the experiments. We train DEGAS with Adam~\cite{KingBa15adam} for 800k iterations.

\begin{figure}
    \centering
    \includegraphics[width=\linewidth]{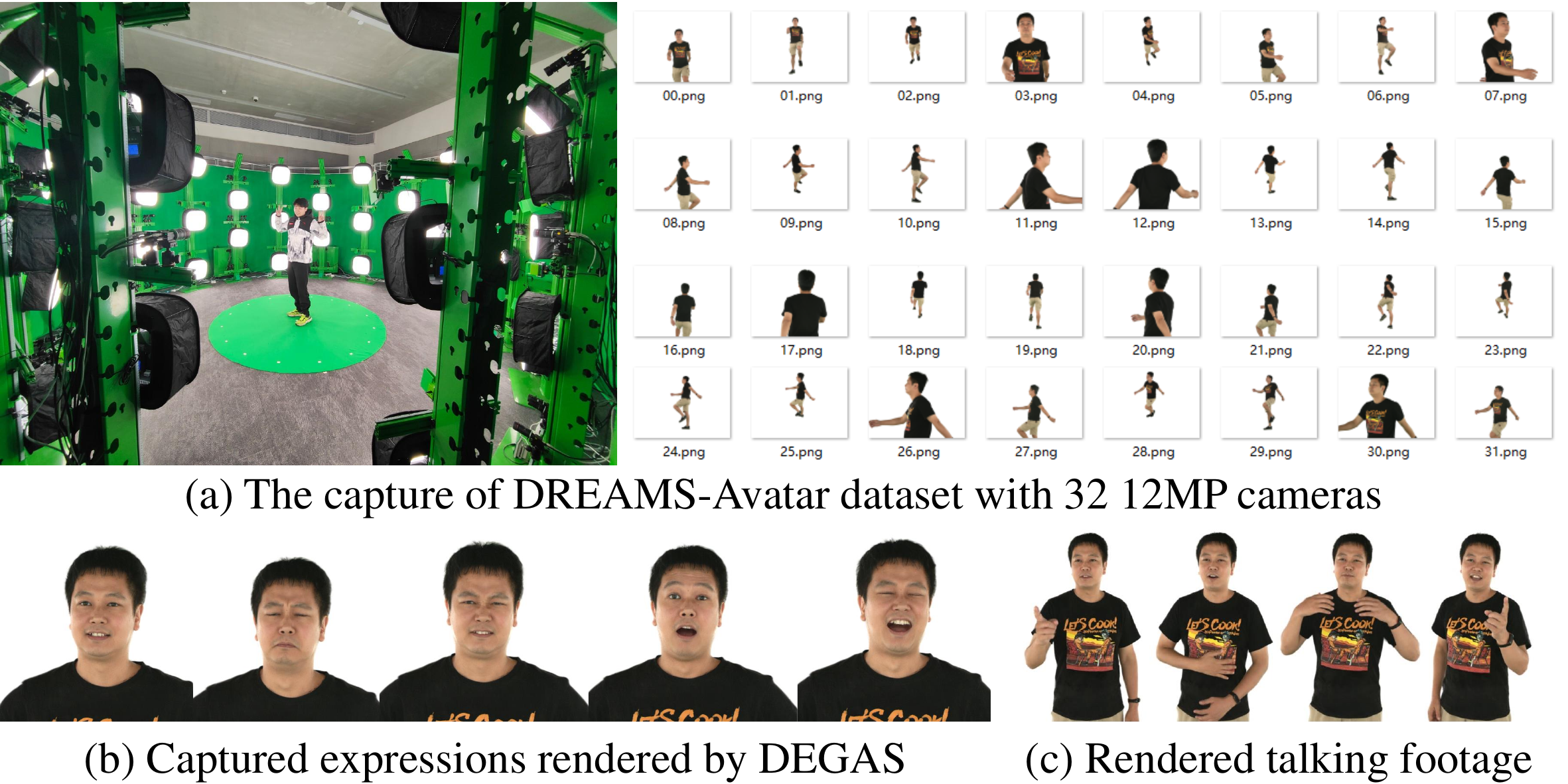}
    \caption{
        \textbf{The DREAMS-Avatar dataset.}
        6 subjects captured with 32 12MP cameras performing large body poses and rich facial expressions, e.g., smiling, laughing, angry, surprised, and a sequence of talking or singing.
    }
    \label{fig:dataset}
\end{figure}

\begin{figure*}[htbp]
    \centering
    \includegraphics[width=\linewidth]{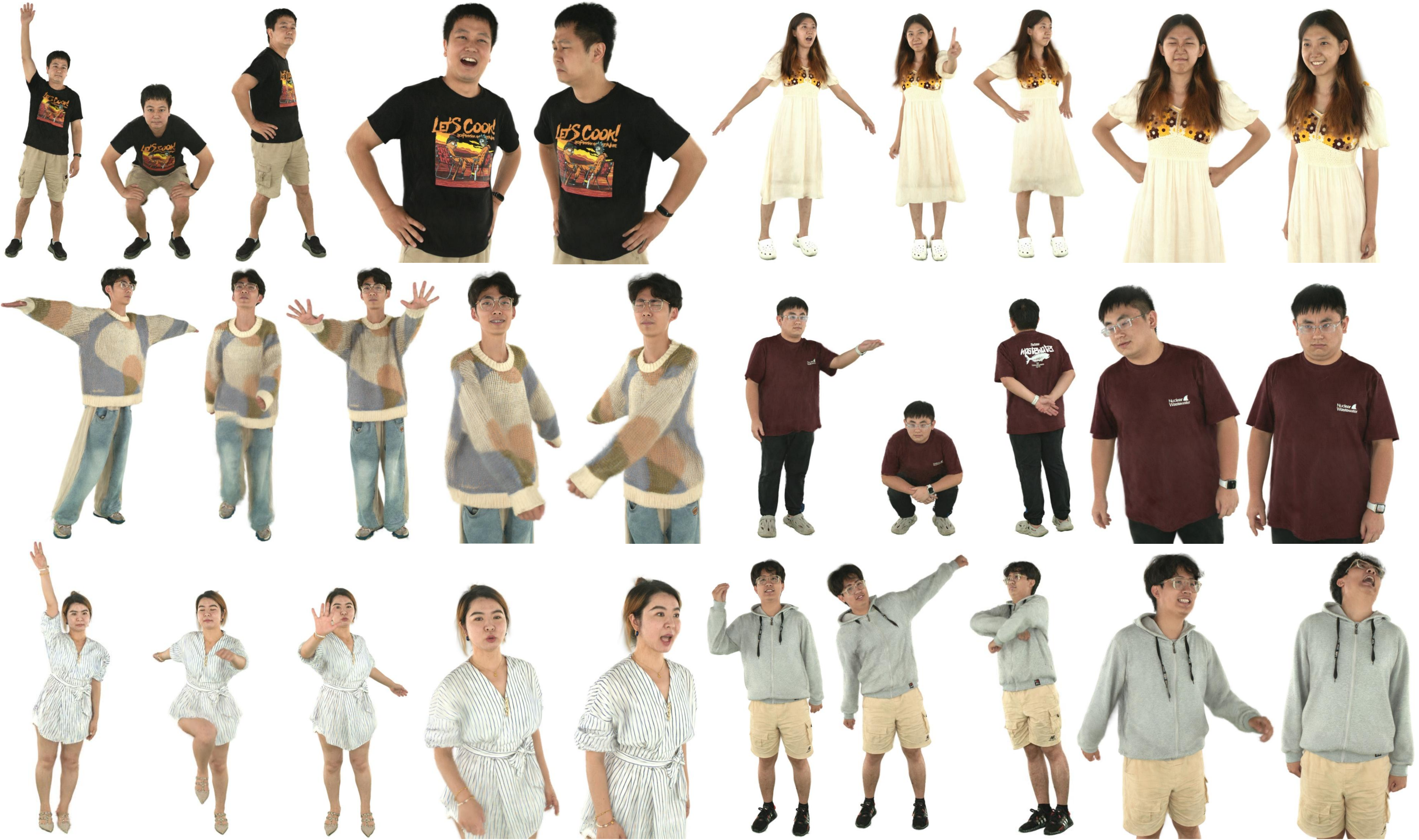}
    \caption{ 
    \textbf{Rendering results of our method.} 
    We show rendering results of our method DEGAS on the DREAMS-Avatar dataset.
    DEGAS is able to render high quality avatars with
    large body pose variations and rich facial expression details.
    }
    \vspace{-1.0em}
    \label{fig:exp-viewsync}
\end{figure*}

\begin{figure}
    \centering
    \includegraphics[width=\linewidth]{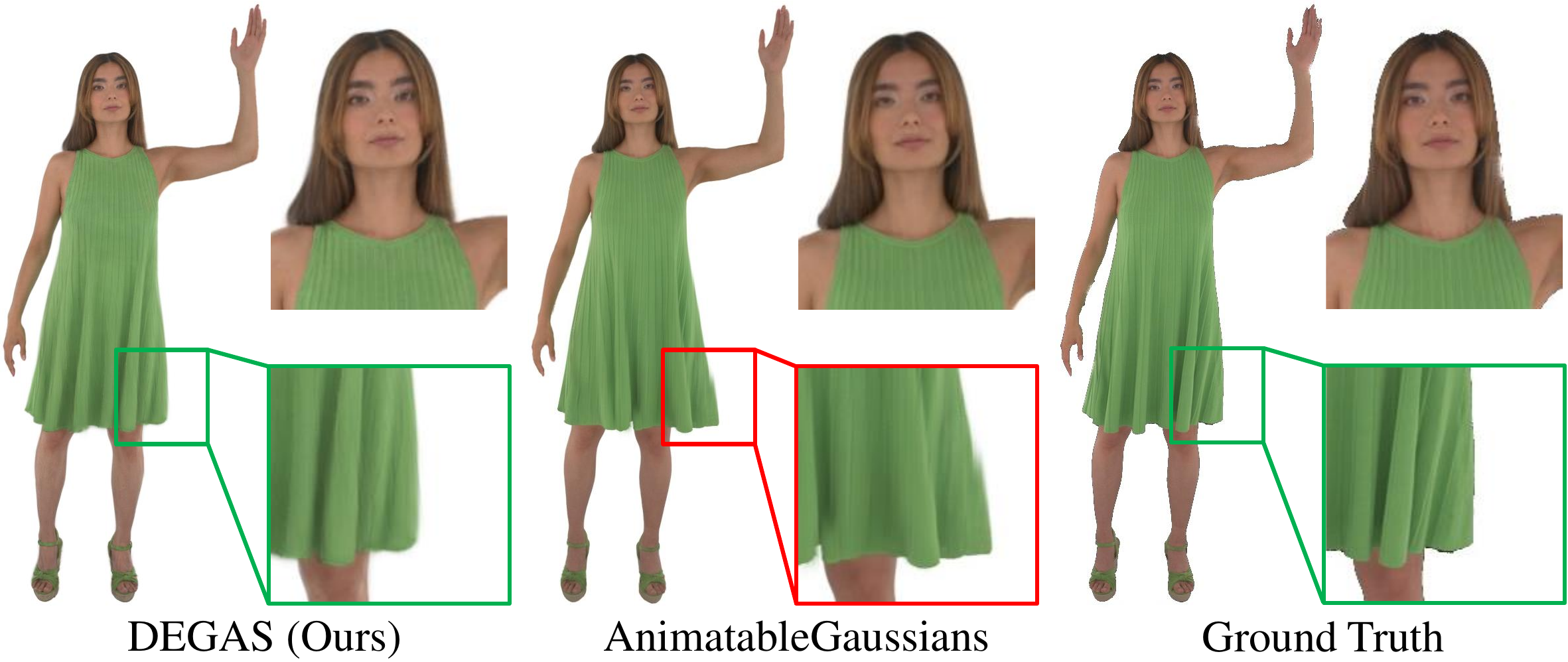}
    \caption{
        \textbf{Comparison on full-body avatars.}
        Our method renders high quality details on ActorsHQ.
    }
    \vspace{-0.5em}
    \label{fig:exp-actorshq}
\end{figure}

\begin{figure}
    \centering
    \includegraphics[width=\linewidth]{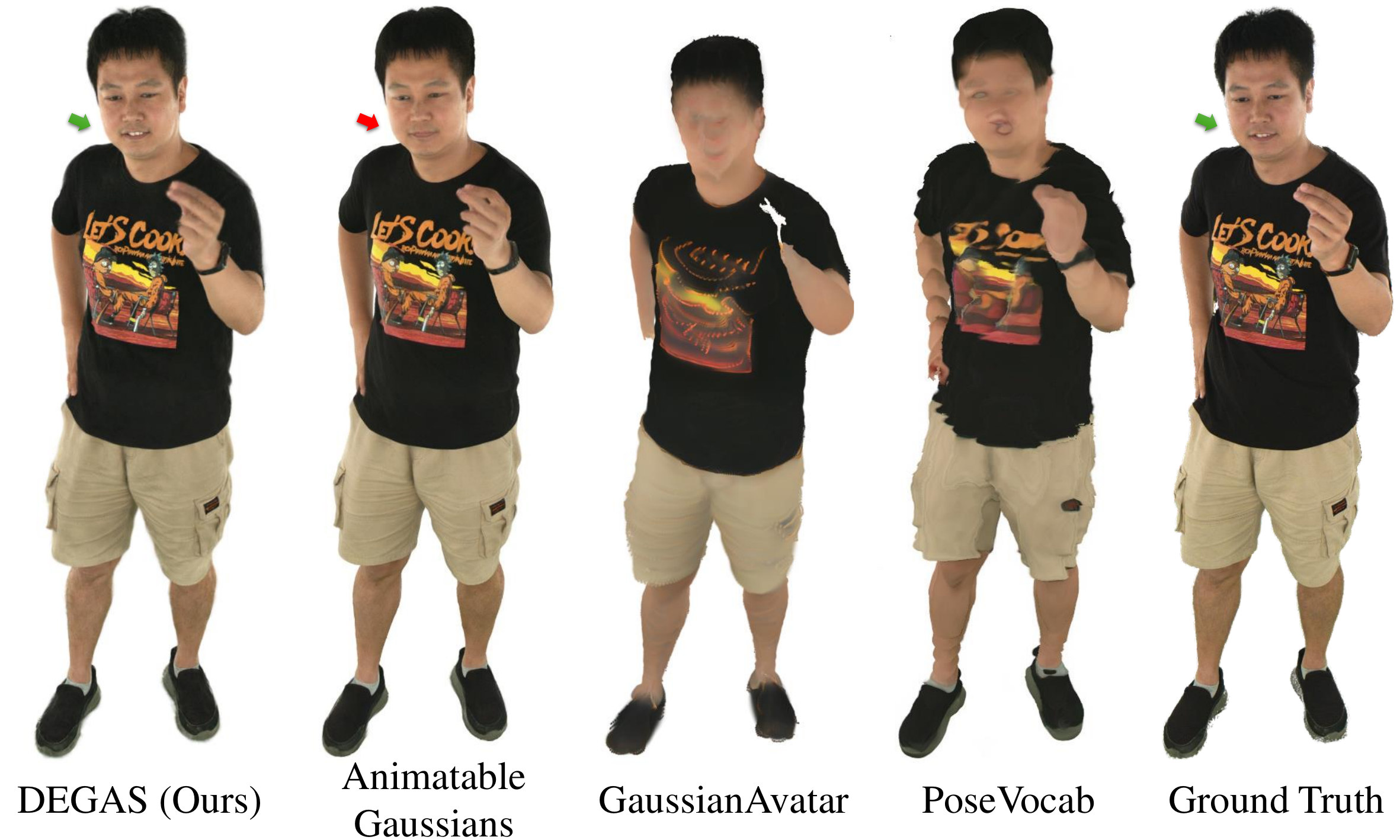}
    \caption{
        \textbf{Comparison on our DREAMS-Avatar dataset.}
        Our method renders high quality details comparing to SoTA methods.
    }
    \vspace{-0.2in}
    \label{fig:exp-dreams-view}
\end{figure}

\section{Experiments}
\label{sec:exp}

\noindent
\textbf{ActorsHQ Dataset}.
ActorsHQ is a high-quality dataset for full-body avatars.
We follow the experiment setup of AnimatableGaussians~\cite{li2024animatablegaussians} to use 47 full-body views (46 views for training and 1 for testing, 
at 1k resolution).

\noindent
\textbf{DREAMS-Avatar Dataset}.
Due to the lack of datasets for evaluating full-body talking avatars with rich facial expressions, we propose the \textit{DREAMS-Avatar} dataset.
DREAMS-Avatar includes the performance of 6 subjects captured with 32 12MP cameras, each with 2 sequences. 
The first of which is the footage of standard poses and facial expressions, while the second is a freestyle talking or singing.
We show in \Cref{fig:teaser,fig:dataset} the large body poses, rich facial expressions, and challenging clothes and glasses in the dataset.
We aim to cover the pose and expression variations in a tele-presentation scenario.

\begin{table}[t]
\centering
\resizebox{\columnwidth}{!}{
\begin{tabular}{lcccc}
    \toprule
    Method & PSNR$\uparrow$ & SSIM$\uparrow$ & LPIPS$\downarrow$ & FID$\downarrow$ \\
    \midrule
    GaussianAvatar~\cite{hu2024gaussianavatar} & 26.9497 & 0.9389 & 0.0407 & 38.5387 \\
    3DGS-Avatar~\cite{qian20233dgsavatar} & 28.7836 & 0.9511 & 0.0418 & 49.3673 \\
    AnimatableGaussians~\cite{li2024animatablegaussians} & \underline{30.3607} & \underline{0.9682} & \underline{0.0339} & \underline{33.4665} \\
    \midrule
    Ours & \textbf{31.1262} & \textbf{0.9708} & \textbf{0.0318} & \textbf{24.4555} \\
    \bottomrule
\end{tabular}
}
\caption{
    \textbf{Quantitative comparison on full-body avatars.}
    Experiments conducted on ActorsHQ Actor01/Sequence1 following the setup of AnimatableGaussians~\cite{li2024animatablegaussians}. Our method quantitatively outperforms all three SoTA methods.
}
\vspace{-0.2in}
\label{tab:exp_actorshq}
\end{table}

\subsection{Comparison on Full-body Avatars}
We conducted experiments on ActorsHQ in comparison to state-of-the-art (SoTA) methods including 3DGS-Avatar~\cite{qian20233dgsavatar}, GaussianAvatar~\cite{hu2024gaussianavatar}, and AnimatableGaussians~\cite{li2024animatablegaussians}.
Different from well-aligned 3DMMs available in a multi-view head avatar dataset~\cite{kirschstein2023nersemble}, registration in a full-body setup is usually compromised by clothing (inaccurate surface alignment), large poses (inaccurate joints), moving head (inaccurate face alignment), etc.

\Cref{fig:exp-actorshq} shows the qualitative comparison on ActorsHQ. 
The rendering results of our method exhibit rich texture details.
The quantitative comparison is reported in \Cref{tab:exp_actorshq}.
We report PSNR and SSIM calculated in the rendered full images, LPIPS~\cite{zhang2018perceptual} and FID~\cite{heusel2017fid} in the cropped regions.
Our method quantitatively outperforms all three SoTA methods.
One key observation we make is that the adoption of powerful 2D CNN networks in both AnimatableGaussians~\cite{li2024animatablegaussians} and ours significantly improves pose-dependent modeling.
More discussions are in \Cref{sec:comp_ag}.

\begin{figure*}[htbp]
    \centering
    \includegraphics[width=\linewidth]{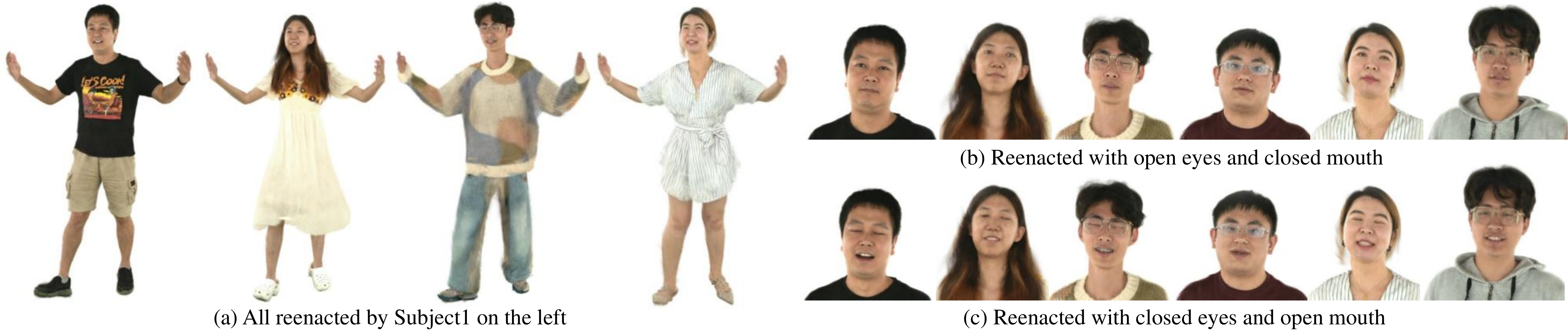}
    \caption{ 
    \textbf{Reenactment results of our method.} 
    We show the rendering results of all subjects reenacted by Subject1's Sequence2.
    Most subjects are reenacted correctly on eyes and mouths except that Subject4's eyes control was affected by the reflections on the glasses.
    }
    \label{fig:exp-reenact}
\end{figure*}

\begin{table}[t]
\centering
\resizebox{\columnwidth}{!}{
\begin{tabular}{lccccc}
    \toprule
    Method & PSNR$\uparrow$ & SSIM$\uparrow$ & LPIPS$\downarrow$ & FID$\downarrow$ & AED$\downarrow$ \\
    \midrule
    AnimatableGaussians~\cite{li2024animatablegaussians} & \underline{32.1534} & \underline{0.9814} & \underline{0.0167} & \textbf{13.0829} & \underline{0.2657} \\
    PoseVocab~\cite{li2023posevocab} & 28.3966 & 0.9740 & 0.0503 & 147.1550 & - \\
    GaussianAvatar~\cite{hu2024gaussianavatar} & 20.9320 & 0.9582 & 0.1056 & 81.0665 & - \\
    \midrule
    DEGAS (ours) & \textbf{33.9613} & \textbf{0.9853} & \textbf{0.01520} & \underline{13.9276} & \textbf{0.0598} \\
    \bottomrule
\end{tabular}
}
\caption{
    \textbf{Quantitative comparison on DREAMS-Avatar.}
    Our method outperforms other SoTA methods in terms of PSNR, SSIM, and LPIPS, and has on-par FID with AnimatableGaussians~\cite{li2024animatablegaussians}. 
    Expression accuracy AED of our method is significantly better. PoseVocab~\cite{li2023posevocab} and GaussianAvatar~\cite{hu2024gaussianavatar} failed to reconstruct the face region in the DREAMS-Avatar dataset.
}
\vspace{-0.05in}
\label{tab:exp_viewsync}
\end{table}

\subsection{Comparison on DREAMS-Avatar}
\label{sec:comp_dreams}
For the experiments conducted on DREAMS-Avatar, we take {Sequence1} of each subject for training with 31 cameras excluding {Cam02}, which is used only for testing. 
We evaluate view synthesis on {Sequence1}, novel pose and facial expression reenactment on {Sequence2}.

\noindent
\textbf{View Synthesis.}
\Cref{tab:exp_viewsync} shows the quantitative results of view synthesis on {Sequence1 Cam02}'s 500--1000 frames where both large poses and rich facial expressions are performed.
The facial expression accuracy is evaluated by the \textit{Average Expression Distance}~(AED) proposed in PIRenderer~\cite{ren2021pirenderer}, which estimates the cosine distance of the expression coefficients extracted by Deep3DFaceRecon~\cite{deng2019deep3dface}.
We show the qualitative comparison in \Cref{fig:exp-dreams-view}.
AnimatableGaussians~\cite{li2024animatablegaussians} can render a high-quality body and a neutral face. PoseVocab~\cite{li2023posevocab} and GaussianAvatar~\cite{hu2024gaussianavatar} fail to model the face region due to the large expression variations in the dataset.
While our method renders high-quality body and face details.
We show more rendering results of DEGAS in \Cref{fig:exp-viewsync} with various poses and rich facial expressions.

\noindent
\textbf{Novel Poses and Expressions.}
We demonstrate the reenactment of DEGAS to novel poses. As shown in \Cref{fig:exp-reenact}, all avatars trained are reenacted by Subject1's Sequence2. Both same-identity and cross-identity reenactments show high-quality details. 
Specifically on the face regions, 
DEGAS responds correctly to eyes and mouth control.



\begin{table}[t]
\centering
\resizebox{\columnwidth}{!}{
\begin{tabular}{lccc}
    \toprule
    \multirow{2}{*}{Method} & \multirow{2}{*}{FPS$\uparrow$} & \multirow{2}{*}{Training Time$\downarrow$} & Disentangled \\
    ~ & ~ & ~ & Encoder/Decoder \\
    \midrule
    AnimatableGaussians~\cite{li2024animatablegaussians} & 10 & 160 hours & $\times$ \\
    DEGAS (ours) & \textbf{30} & 55 hours & $\checkmark$ \\
    \bottomrule
\end{tabular}
}
\caption{
    \textbf{Comparison with AnimatableGaussians~\cite{li2024animatablegaussians}.}
    Our method reaches real-time framerate and favors disentangled encoder and decoder which contribute to potential applications. Numbers recorded on one NVIDIA RTX3090 and training with 800k iterations according to the original paper of AnimatableGaussians~\cite{li2024animatablegaussians}.
}
\vspace{-0.05in}
\label{tab:comp_ag}
\end{table}

\subsection{Discussion w.r.t. AnimatableGaussians}
\label{sec:comp_ag}
Both our method and AnimatableGaussians~\cite{li2024animatablegaussians} learn from CodecAvatars~\cite{bagautdinov2021drivingsignal} in adopting powerful 2D CNN networks for the generation of Gaussian maps.
The difference is that ours does not use skip connections between encoder and decoder, considering the potential streaming scenario~\cite{pixel_codec_avatars}.
Also, the bottleneck between the encoder and decoder that follows the UV layout of SMPL-X enables the face signal injection in our method.
The runtime is compared in \Cref{tab:comp_ag} where our method reaches a real-time framerate.

\subsection{Ablation Study}
\label{sec:ablation}

\begin{table}[t]
\centering
\resizebox{\columnwidth}{!}{
\begin{tabular}{lccccc}
    \toprule
    ~ & \multicolumn{4}{c}{Same-Identity Reenactment} & Cross-Identity \\
    ~ & PSNR$\uparrow$ & SSIM$\uparrow$ & LPIPS$\downarrow$ & AED$\downarrow$ & AED$\downarrow$ \\
    \midrule
    w/ DECA & 26.185 & 0.954 & 0.049 & 0.1105 & 0.2638 \\
    w/ DAD-3DHeads & 26.233 & 0.954 & 0.049 & 0.0998 & 0.3697 \\
    w/ DPE (ours) & {26.212} & {0.954} & {0.049} & \textbf{0.0853} & \textbf{0.2413} \\
    \bottomrule
\end{tabular}
}
\caption{
    \textbf{Choice of face signal.}
    The 2D talking faces-based expression encoder from DPE~\cite{DPE:CVPR:2023} is a better choice as face signals.
}
\vspace{-0.1in}
\label{tab:exp_face}
\end{table}

\begin{figure}
    \centering
    \includegraphics[width=\linewidth]{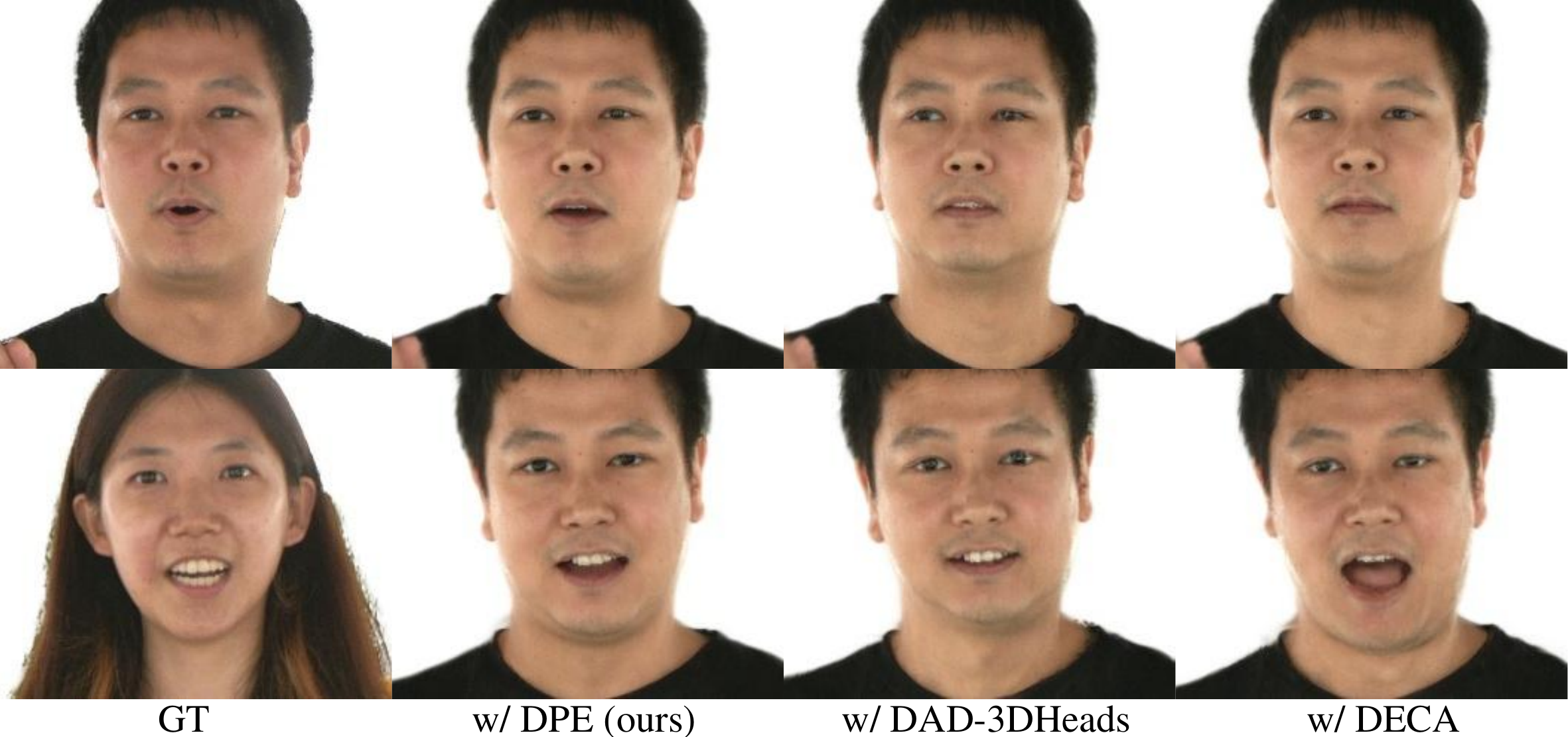}
    \caption{
        \textbf{Ablation on facial reenactment.}
        Using expression encoder from DPE~\cite{DPE:CVPR:2023} drives more similar expressions comparing to using DECA~\cite{DECA:Siggraph2021} or DAD-3DHeads~\cite{dad3dheads:CVPR:2022}.
    }
    \label{fig:exp_face}
\end{figure}

\noindent
\textbf{Choice of face signal.}
We conducted ablation experiments to validate the design choice of using a pretrained expression encoder from DPE~\cite{DPE:CVPR:2023} as the face signal.
We compared to using a 3DMM as the face signal 
by extracting FLAME parameters from the frontal camera.
We replaced the \textit{jaw pose} of SMPL-X with that from the extracted FLAME parameters and used the \textit{expression} parameters to condition our convolutional decoder.
Note that in our method, the \textit{jaw pose} is neutralized, i.e., the jaw movements are solely represented by the cVAE.

We tested two approaches for extracting FLAME parameters, i.e., DECA~\cite{DECA:Siggraph2021} and DAD-3DHeads~\cite{dad3dheads:CVPR:2022}.
For same-identity reenactment, we trained on {Subject1} {Sequence1} and tested on {Sequence2}. For cross-identity reenactment, we reenacted the avatar by the motion of {Subject2} {Sequence2}.
As shown in \Cref{fig:exp_face}, the 2D talking faces-based expression encoder from DPE~\cite{DPE:CVPR:2023} enables more similar expressions. 
\Cref{tab:exp_face} shows the quantitative comparison. 

\begin{table}[t]
\centering
{
\begin{tabular}{lcccc}
    \toprule
    ~ & PSNR$\uparrow$ & SSIM$\uparrow$ & LPIPS$\downarrow$ & FID$\downarrow$ \\
    \midrule
    3 views & 29.6554 & 0.9748 & 0.0262 & {33.4759} \\
    6 views & {30.9780} & {0.9782} & {0.0188} & {36.6235} \\
    12 views & {30.9518} & {0.9783} & {0.0189} & {22.7458} \\
    \midrule
    31 views & {32.6349} & {0.9823} & {0.0166} & {14.2061} \\
    \bottomrule
\end{tabular}
}
\caption{
    \textbf{Training with sparse views.}
    Our method can be trained with sparse views with decent quality.
}
\vspace{-0.05in}
\label{tab:ablation_views}
\end{table}

\noindent
\textbf{Training with Sparse Views.}
Our avatar modeling method is trained in a multi-view setup. We report in \Cref{tab:ablation_views}
that DEGAS can be trained with 3 views, 6 views, and 12 views.

\begin{table}[t]
\centering
\resizebox{\columnwidth}{!}{
\begin{tabular}{lcccc}
    \toprule
    ~ & PSNR$\uparrow$ & SSIM$\uparrow$ & LPIPS$\downarrow$ & FID$\downarrow$ \\
    \midrule
    w/o Pose $\boldsymbol\theta$ Embedding & 30.0084 & 0.9695 & 0.0289 & \underline{23.8262} \\
    w/o PVM Encoder & \underline{30.4173} & \underline{0.9729} & \underline{0.0284} & 24.2348 \\
    w/o PVE Encoder & 29.4263 & 0.9656 & 0.0338 & 26.5531 \\
    Ours & \textbf{30.6700} & \textbf{0.9731} & \textbf{0.0281} & \textbf{23.8110} \\
    \bottomrule
\end{tabular}
}
\caption{
    \textbf{Ablation on encoder branches.}
    All three encoder branches help improve the rendering quality of our method.
}
\vspace{-0.05in}
\label{tab:exp_encoder}
\end{table}

\noindent
\textbf{Pose Encoders.}
There are three pose encoders in our method to generate body signals. 
We conducted ablation studies on ActorsHQ Actor02 by disabling one of them each time. 
The quantitative results are reported in \Cref{tab:exp_encoder}.

\begin{figure}
    \centering
    \includegraphics[width=\linewidth]{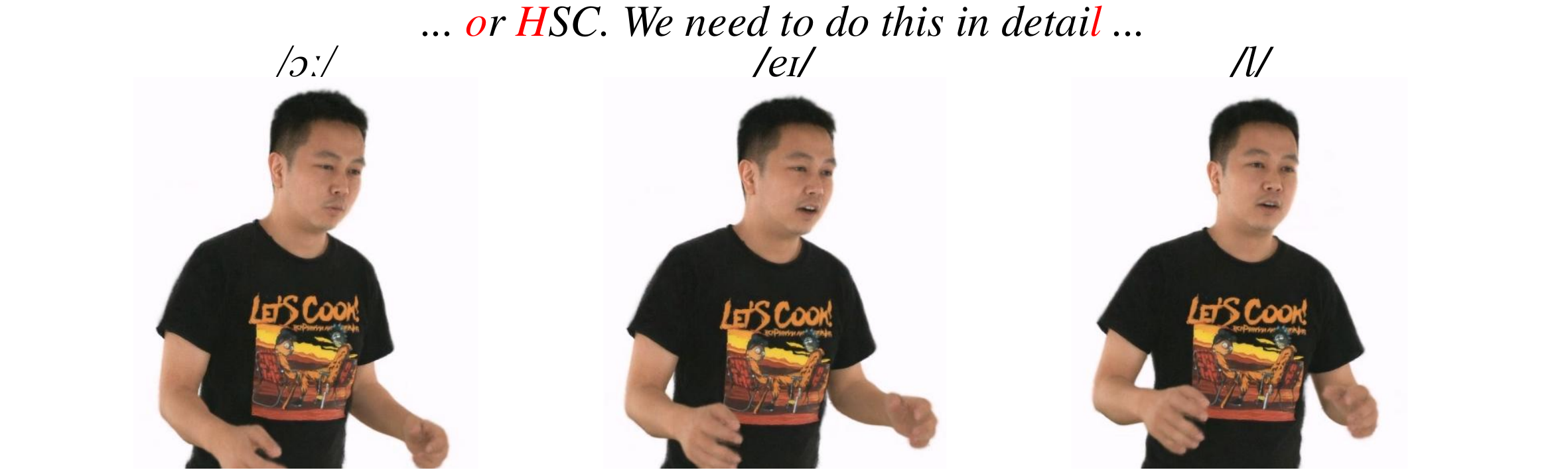}
    \caption{
        \textbf{Audio-driven example of our method.}
        Given an audio sequence, we generate face signal from SadTalker~\cite{Sadtalker:CVPR:2023} and DPE~\cite{DPE:CVPR:2023}, and body signal from TalkSHOW~\cite{yi2022talkshow}.
    }
    \vspace{-0.1in}
    \label{fig:exp_audio}
\end{figure}

\subsection{3D Full-body Talking Avatars}
The use of a pretrained encoder from 2D talking faces lets our method inherit the ability to both video-driven and audio-driven reenactment.
With the help of SadTalker~\cite{Sadtalker:CVPR:2023}, we show the extension of DEGAS to full-body talking avatars.
Given an audio clip,
we firstly use SadTalker~\cite{Sadtalker:CVPR:2023} to generate 2D talking videos from one face image of the subject,
and then use the DPE~\cite{DPE:CVPR:2023} expression encoder to extract face signal to drive our avatars.
The body pose generation is a whole another research track~\cite{yi2022talkshow, ng2024audio2photoreal, chhatre2024amuse}. 
Here we only showcase the generation from TalkSHOW~\cite{yi2022talkshow} in \Cref{fig:exp_audio}. Noted that the \textit{jaw pose} generated by TalkSHOW is not used.

\subsection{Limitations}
The pretrained expression encoder from DPE~\cite{DPE:CVPR:2023} that we explore in this paper has the advantage of being trained with a large collection of face images.
Yet the quality of the 2D talking faces method itself limits the reenactment quality of our method.
We do observe pose and identity-related information being not fully disentangled from the expression.
We believe that using encoders from more advanced 2D talking faces methods~\cite{xu2024vasa} would help.
Another issue is that the clothes are not modeled in a separate layer, causing artifacts for the loose cloth in challenging poses.

\section{Conclusion}
In this paper, we proposed DEGAS, the first 3DGS-based method for full-body avatars with subtle and accurate facial expressions.
We discussed in this paper that an expression latent space pretrained solely on 2D talking faces, 
is a better choice for the reenactment of 3D avatars, 
opening new possibilities for interactive life-like agents.
The avatars modeled with DEGAS can be animated and rendered in real-time framerate to perform natural body motions and rich facial expressions.
We conducted qualitative and quantitative experiments to validate the efficacy of our method.
We also showed the audio-driven extension of our method to demonstrate its potential for downstream applications.

\section{Acknowledgments}
We thank the HKUST(GZ) DREAMS Lab for assisting in data collection.
The work is supported by the National Social Science Foundation of China under grant number 24VWB020.

{
    \small
    \bibliographystyle{ieeenat_fullname}
    \bibliography{main}
}
\clearpage

\renewcommand{\thefigure}{A\arabic{figure}}
\setcounter{figure}{0}
\renewcommand{\thetable}{A\arabic{table}}
\setcounter{table}{0}
\renewcommand{\thesection}{\Alph{section}}
\setcounter{section}{0}

\twocolumn[{
\begin{center}
    \Large
    \textbf{\thetitle}\\
    \vspace{0.5em}Supplementary Material \\
    \vspace{1.0em}
    
    \captionsetup{type=figure}
    \includegraphics[width=\linewidth]{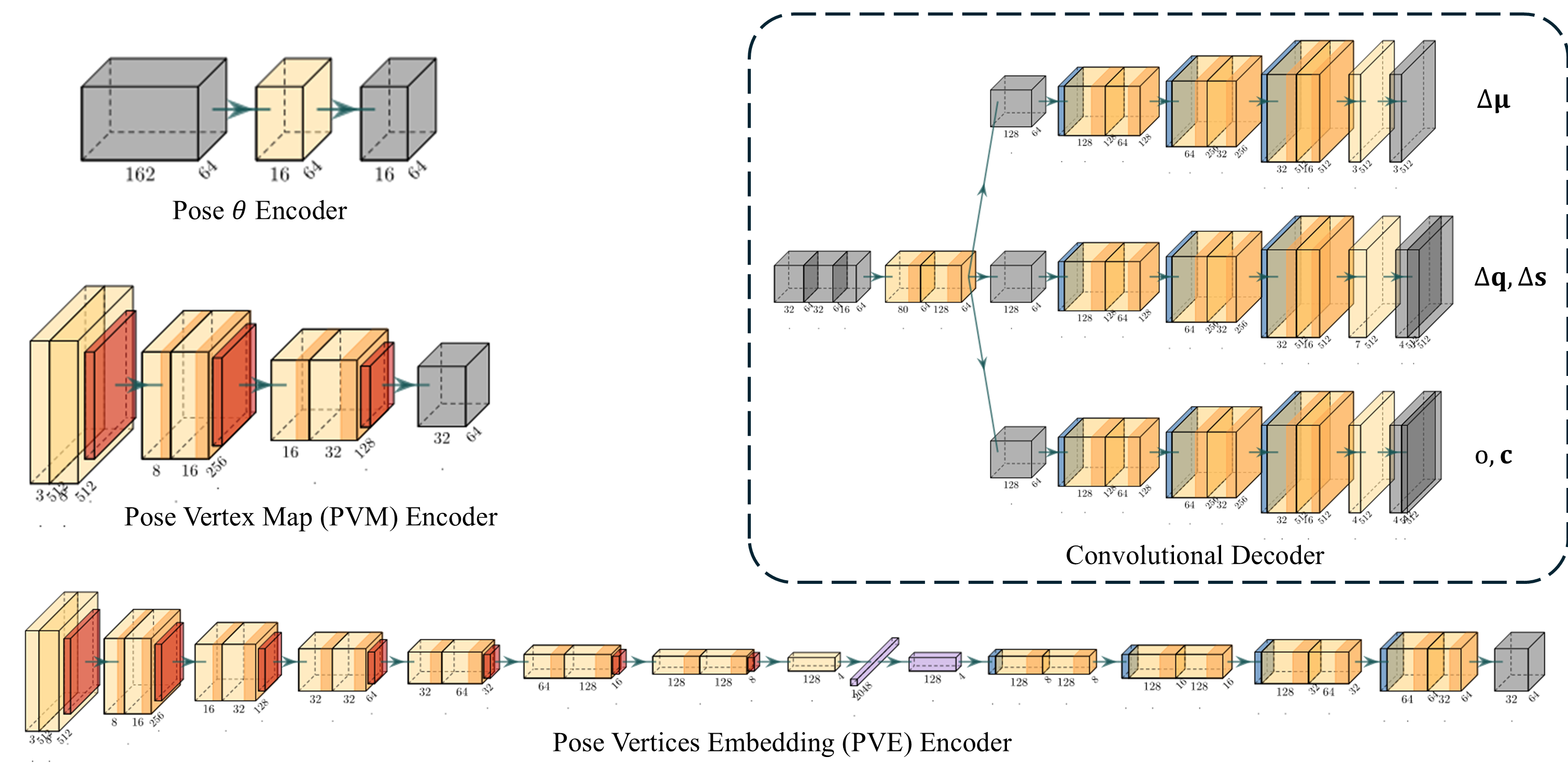}
    \captionof{figure}{
    \textbf{Network structures.} 
    We show the network structures of the three encoder branches and the convolutional decoder.
    }
    \label{figa:suppl_net}
\end{center}
}]

\begin{figure}
    \centering
    \includegraphics[width=\linewidth]{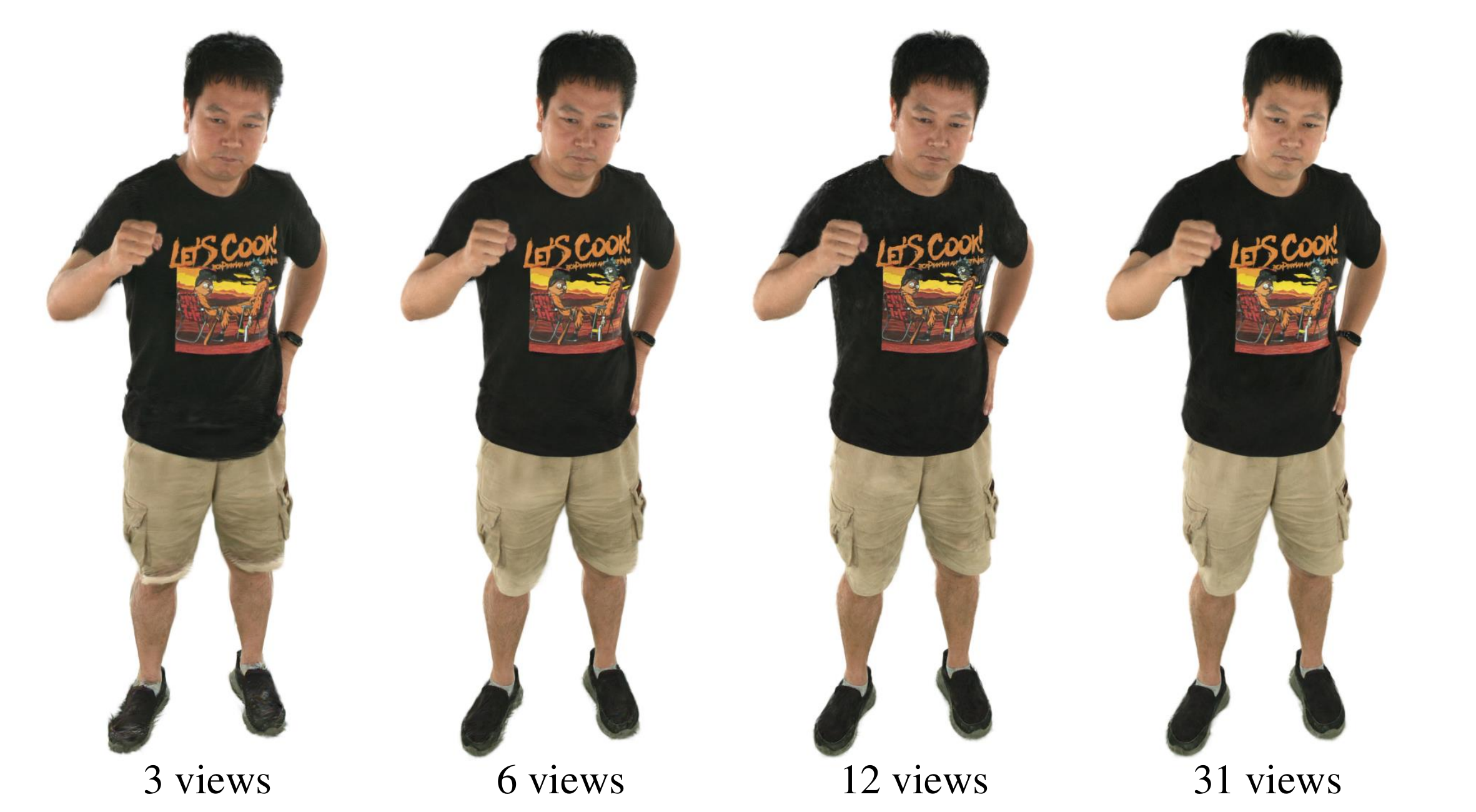}
    \caption{
        \textbf{Ablation on sparse views.}
        Our method can be trained with sparse views, i.e., 3 views, 6 views, and 12 views.
    }
    \label{figa:ablation_views}
\end{figure}

\section{Implementation Details}
\label{sec:suppl_impl}
There are three encoder branches and one convolutional decoder defined in this paper. We show the network structures in \Cref{figa:suppl_net}. 
As can be seen that though both the encoder PVM and PVE work on the position map, PVE is much deeper than PVM. 
We find all three encoder branches help the modeling.
The convolutional decoder has three groups of CNNs, i.e., one for position, one for rotation and scaling, and the last one for opacity and color.

\section{Sparse Views}
We show in \Cref{figa:ablation_views} that our method can be trained with sparse views.

\section{Ethics}
\label{sec:ethics}
We captured six human subjects in our proposed DREAMS-Avatar dataset.
All subjects have given written consent for the captured footage in this dataset.
We make the data publicly available for research purposes.

Our method could be extended to synthesize communication media contents of real people.
We do not condone using our work to generate fake images or videos of any person with the intent of spreading misinformation or tarnishing their reputation.

\end{document}